\documentclass[10pt,twocolumn,letterpaper]{article}

\usepackage[pagenumbers]{cvpr} % To force page numbers, e.g. for an arXiv version

\usepackage{enumitem}
\usepackage{xspace}
\usepackage{ifthen}
\usepackage[normalem]{ulem}

\newcommand{\SB}[2][]{%
    \ifthenelse{ \equal{#1}{} }
        {\textcolor{orange}{#2\xspace{}}}
        {\textcolor{orange}{\sout{#1} #2\xspace{}}}
}

\definecolor{cvprblue}{rgb}{0.21,0.49,0.74}
\usepackage[pagebackref,breaklinks,colorlinks,allcolors=cvprblue]{hyperref}

\usepackage{tabularx}
\usepackage{booktabs}
\usepackage{array}

\newcolumntype{Y}{>{\centering\arraybackslash}X}

\def\paperID{8} % *** Enter the Paper ID here
\def\confName{CVPR}
\def\confYear{2026}

\title{Cross-Modal Knowledge Distillation from Spatial Transcriptomics to Histology}

\author{
    Arbel Hizmi$^{1,2}$ \quad Artemii Bakulin$^1$ \quad Shai Bagon$^1$ \quad Nir Yosef$^1$ \\
    $^1$Weizmann Institute of Science \quad $^2$Reichman University \\
    {\tt \{arbel.hizmi, artemii.bakulin, shai.bagon, nir.yosef\}@weizmann.ac.il} \\
    {\small \url{https://cross-modal-distillation.github.io/}}
}

\begin{document}
% teaser figure -- code MUST be here!

\makeatletter
\twocolumn[{%
\begin{@twocolumnfalse}
\maketitle

\begin{center}
  \phantomsection

  \captionsetup{type=figure,hypcap=false}
  \includegraphics[width=1.0\linewidth]{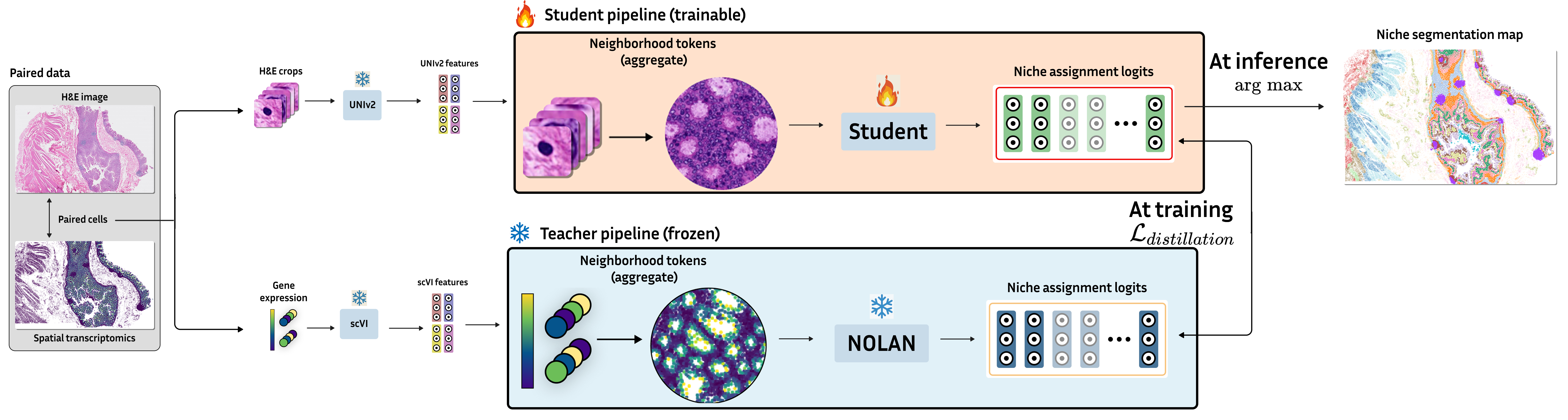}

  \caption{%
  \textbf{Framework overview}: Our goal is to transfer spatially resolved molecular niche structure from spatial transcriptomics to histology, so that niche organization can be predicted from H\&E alone at inference time. The key idea is that transcriptomics provides a richer view of local tissue state, while histology is abundant and widely available. During training, we therefore use a frozen spatial-transcriptomics teacher to supervise a histology student on matched cell neighborhoods. For each cell, both teacher and student operate on spatial neighborhoods: the student takes UNIv2 features from H\&E cell crops, while the teacher takes scVI features from the paired transcriptomic measurements. The frozen pretrained NOLAN teacher produces niche-assignment logits, and the student is trained with a distillation loss to match them. At inference, only the histology branch is used, and the student predicts niche labels from H\&E neighborhoods alone. Flame symbol: ``trainable''; Snowflake symbol: ``frozen''.%
  }
  \label{fig:overview}
\end{center}

\end{@twocolumnfalse}
}]
\makeatother

\begin{abstract}
Spatial transcriptomics provides a molecularly rich description of tissue organization, enabling unsupervised discovery of tissue niches --- spatially coherent regions of distinct cell-type composition and function that are relevant to both biological research and clinical interpretation. However, spatial transcriptomics remains costly and scarce, while H\&E histology is abundant but carries a less granular signal. We propose to leverage paired spatial transcriptomics and H\&E data to transfer transcriptomics-derived niche structure to a histology-only model via cross-modal distillation. Across multiple tissue types and disease contexts, the distilled model achieves substantially higher agreement with transcriptomics-derived niche structure than unsupervised morphology-based baselines trained on identical image features, and recovers biologically meaningful neighborhood composition as confirmed by cell-type analysis. The resulting framework leverages paired spatial transcriptomic and H\&E data during training, and can then be applied to held-out tissue regions using histology alone, without any transcriptomic input at inference time.
\end{abstract}    
\section{Introduction}
\label{sec:intro}

Many tissue phenomena — such as immune infiltration, stromal remodeling, and tumor progression — are not properties of isolated cells, but emerge from the spatial organization and interactions of multiple cell types. A central goal in tissue analysis is therefore to identify spatially coherent regions that reflect characteristic local cell compositions and interactions, commonly referred to as \textit{cellular niches}. The task of partitioning tissue into such regions is known as \textit{tissue niche segmentation}.

While traditional approaches to tissue structure analysis rely on manual annotation, recent work has shown that spatial transcriptomics provides a rich substrate for unsupervised niche discovery. Spatial transcriptomics jointly measures gene expression and spatial coordinates across thousands of cells, providing a high-dimensional molecular fingerprint of the tissue microenvironment. Methods such as BANKSY, SpaGCN, MENDER, and NOLAN \cite{singhal2024banksy, hu2021spagcn, Yuan2024Mender, bakulin2025nolan} exploit this richness to define niches directly in molecular feature space, incorporating spatial neighborhood structure to produce coherent, biologically interpretable tissue partitions. However, spatial transcriptomics assays remain costly and are often available only for a subset of samples or tissue regions, limiting their applicability at scale.

In contrast, hematoxylin and eosin (H\&E) histology is universally available across both research and clinical workflows, with archives spanning millions of slides. Yet H\&E does not directly measure gene expression, and morphological features alone may not faithfully recover molecularly defined tissue organization. Recent foundation models have dramatically improved H\&E representation learning \cite{chen2024uni, xu2024wholeslide, vorontsov2024clinicalgrade, hoptimus0}, but even rich morphological embeddings are not guaranteed to capture the molecular distinctions that define transcriptomic niches — particularly in regions where cell-type differences are subtle or where niche identity is determined by microenvironmental context rather than cell morphology alone.

We address the gap between abundant H\&E imaging and the molecular information available in spatial transcriptomics as a cross-modal distillation problem. When paired spatial transcriptomics and H\&E data are available for the same cells, spatial transcriptomics provides a richer molecular view of cell state and neighborhood composition than histology alone. We use the spatial transcriptomics modality to define a reference niche structure with a transcriptomic teacher, and train a histology-based student to match the teacher’s niche predictions from cell-level image embeddings and their spatial neighborhoods. This encourages the student to learn morphology signals that are informative of transcriptomic niches. At inference time, the student uses histology features only, enabling niche prediction without molecular measurements.

We evaluate (i) how well the histology student reproduces the spatial transcriptomics-derived reference structure and (ii) whether the predicted niches preserve biologically meaningful organization, including cell-type composition patterns and (iii) correspondence of predicted niches with pathology annotations of tissue areas.
\section{Related work}
\label{sec:related_work}

\subsection{Spatial transcriptomics and cellular niches}

Many important tissue functions, such as tumor growth and immune infiltration, are governed by the spatial arrangement of cells rather than the characteristics of individual cells. Several methods have been proposed to identify spatial domains or cellular niches by combining gene expression embeddings with spatial relationships \cite{singhal2024banksy, hu2021spagcn, Yuan2024Mender}. A common implicit assumption in existing methods is that tissue can be partitioned into discrete regions. However, Bhate \etal \cite{bhate2022tissue} show that tissue organization is often more complex: tissue regions may lack sharp boundaries, and tissue properties can change gradually across space. As a result, precise niche labels are difficult to define and are often ambiguous. Bakulin \etal \cite{bakulin2025nolan} further argue that manual annotations can be low-resolution or inaccurate under continuous spatial gradients, motivating unsupervised methods.

Given the ambiguity of manual niche annotations, we use NOLAN \cite{bakulin2025nolan} as our transcriptomics teacher. NOLAN is a self-supervised model that learns niche-aware representations from spatial neighborhoods, enabling the discovery of spatially coherent molecular niches without manual annotations. Because NOLAN computes representations from neighborhood-based inputs, the resulting embeddings depend on local cellular context rather than only on individual-cell features. This property is important for our setting, where niches are defined by local spatial context. Accordingly, our method follows NOLAN’s neighborhood-based formulation by using local spatial neighborhoods to learn niche-related representations.

\subsection{Computational pathology}

Deep learning has become central to computational pathology, with recent advances increasingly driven by self-supervised learning and foundation models for H\&E histology. In particular, self-supervised objectives trained on large collections of unlabeled histology images have enabled the development of domain-specific histopathology encoders \cite{chen2024uni, xu2024wholeslide, vorontsov2024clinicalgrade, hoptimus0}. These models learn transferable visual representations that capture tissue morphology across organs and staining variability, and serve as strong backbones for a broad range of downstream tasks. Prior work often evaluates such representations on tasks such as classification and survival prediction. In this work, we use UNIv2 \cite{chen2024uni} as the histology feature encoder. UNIv2 provides general-purpose histology embeddings with strong performance across diverse clinical tasks \cite{chen2024uni}. However, while histology captures rich morphological structure, it does not directly measure molecular state; consequently, tissue niche segmentation from histology alone may not align with molecularly defined organization. To address this gap, we use a transcriptomics teacher and a histology student (initialized from UNIv2), and transfer molecular information into the histology representation through knowledge distillation.

\subsection{Knowledge distillation}

Knowledge distillation is a general framework for transferring information from one model to another by matching softened output distributions \cite{hinton2015distilling}. While knowledge distillation was originally popularized for model compression \cite{Bucilua2006}, it has evolved into a broader family of transfer objectives that distill not only logits but also intermediate representations \cite{Romeroetal2015} and relational structure through contrastive learning \cite{Tianetal2020} and self-distillation in self-supervised learning \cite{caron2021emerging, oquab2023dinov2}. In our work, distillation is used to transfer information between modalities. The teacher operates on transcriptomic embeddings, while the student operates on histology embeddings. By matching niche-assignment logits rather than hard labels, the student is encouraged to capture relative niche similarities and structure encoded by the teacher.

\subsection{Cross-modal representation learning}

Learning unified representations across modalities has become a central paradigm in multimodal deep learning, with landmark works such as CLIP \cite{radford2021learning} demonstrating that aligned embedding spaces enable robust cross-modal transfer.
While most cross-modal approaches in computational pathology target gene expression prediction \cite{Huang2025, hescape, Schmauch2020, Xie2023} or global multimodal alignment \cite{tangle, omiclip}, we focus on transferring \emph{spatial niche structure} — a neighborhood-level property encoding tissue organization rather than individual molecular measurements. Compared to per-gene prediction, \emph{spatial niche structure} represents a more tractable target: it abstracts away molecular noise into a compact summary of tissue organization, while remaining biologically meaningful and naturally aligned with the goal of modeling spatial context.

Accordingly, we treat transcriptomics as a teacher that provides neighborhood-conditioned niche supervision, and histology as a student that learns to approximate this structure from morphology and spatial context. Because ground-truth niche labels are typically unavailable or unreliable, we use spatial transcriptomics-derived niche structure as a higher-information reference. Thus, rather than reproducing an arbitrary teacher partition, we aim to transfer biologically grounded molecular information into histology-only model and to quantify the cross-modal agreement.
\section{Preliminaries}
\label{sec:preliminaries}

\subsection{Neighborhood construction} \label{subsec:neighborhood_construction}

To encode spatial structure, we associate each cell with a local neighborhood defined in physical space (\cref{fig:input_example}, top). For a cell $i$, its neighborhood consists of all cells whose centroids lie within a radius $r$ of the centroid of $i$. A fixed physical radius is not appropriate across datasets, because tissue density varies across assays: the same radius may contain many cells in dense tissues but only a few in sparse tissues, leading to inconsistent amounts of contextual information. To make neighborhood context comparable across datasets and assays, we therefore choose a dataset-specific radius. Specifically, for each dataset we select $r$ such that the expected number of neighbors is approximately equal to a fixed target, estimated empirically by sampling at random locations across the slide. As a result, the radius is larger in sparse tissues and smaller in dense tissues, yielding neighborhoods with a comparable amount of spatial context despite differences in physical cell density.

\subsection{Using spatial transcriptomics to learn tissue structure} \label{subsec:nolan}

NOLAN \cite{bakulin2025nolan} is a self-supervised framework designed to discover spatial cellular niches from spatial transcriptomics data. To encode spatial relationships within each neighborhood, NOLAN uses frequency-based positional encoding computed from relative spatial coordinates. It takes spatial neighborhoods as input and learns niche-aware embeddings in a fully unsupervised manner, assigning each cell to a niche via the activation of its projection head. The resulting niches are evaluated using biologically motivated criteria such as cell-type consistency within niches and correspondence to tissue annotations \cite{bakulin2025nolan}. However, NOLAN operates in gene expression space, and applying it directly to histology images does not guarantee to yield the same quality of results.
\section{Method}
\label{sec:method}

\begin{table*}[t]
\centering
\caption{Agreement with the spatial-transcriptomics teacher niche assignments. We report mean $\pm$ standard deviation of ARI and NMI, computed between each histology-based method and the teacher across all datasets, for $K \in \{10,20\}$ niche categories. ``Ours'' denotes the distilled histology student. Higher values indicate stronger agreement with the teacher-defined niche partition; best results are in \textbf{bold}.}
\label{tab:clustering_performance}
\small
\begin{tabularx}{\textwidth}{l *{4}{>{\centering\arraybackslash}X}}
\toprule
       & \multicolumn{2}{c}{$K = 10$} & \multicolumn{2}{c}{$K = 20$} \\
\cmidrule(lr){2-3}\cmidrule(lr){4-5}
Method & ARI & NMI & ARI & NMI \\
\midrule
Histology NOLAN  & $0.283\pm0.12$ & $0.383\pm0.10$ & $0.234\pm0.08$ & $0.403\pm0.09$ \\
Histology Leiden & $0.191\pm0.12$ & $0.288\pm0.13$ & $0.176\pm0.08$ & $0.327\pm0.11$ \\
\midrule
Ours & \boldmath{$0.615\pm0.11$} & \boldmath{$0.603\pm0.10$} & \boldmath{$0.500\pm0.10$} & \boldmath{$0.579\pm0.08$} \\
\bottomrule
\end{tabularx}
\end{table*}

Our goal is to predict tissue niches from H\&E alone at inference time, while leveraging the richer niche structure available in spatial transcriptomics during training. NOLAN learns niche assignments from spatial transcriptomics neighborhoods and transcriptomic signals that are not directly observed in H\&E images. This motivates transferring niche information from a transcriptomic teacher, rather than relying only on a direct histology adaptation. Accordingly, we use NOLAN as the transcriptomic teacher and train an H\&E-based student to match its niche-assignment logits on paired H\&E-transcriptomics data (\cref{fig:overview}).

For each cell, we extract an H\&E crop centered at its location and encode it with a frozen UNIv2 backbone \cite{chen2024uni} to obtain a cell embedding. We then construct a spatial neighborhood around each cell as described in \cref{subsec:neighborhood_construction}, add the relative positional encoding used by NOLAN (\cref{subsec:nolan}), and feed the neighborhood tokens to a student transformer that outputs a $K$-dimensional vector of niche-assignment logits. During training, we encode the paired transcriptomics with a frozen scVI encoder \cite{lopez2018deep}, construct neighborhoods in the same way, and pass the resulting tokens to a frozen pretrained NOLAN teacher to obtain target logits. The student is trained with a distillation loss that aligns its logits with those of the teacher. At inference, only H\&E is required: the student predicts per-cell niche logits, and, following NOLAN, the niche with the highest logit is taken as the discrete niche label. These per-cell labels can then be visualized as a tissue-level niche map.

\subsection{Spatial transcriptomic-based supervision}

Niche structure learned from spatial transcriptomics provides a useful reference for supervising histology models in the absence of reliable niche annotations. Therefore, we adopt NOLAN model \cite{bakulin2025nolan} as the transcriptomics teacher in our cross-modal distillation method. For each cell, gene expression is first encoded using a frozen scVI model \cite{lopez2018deep} to obtain latent embeddings. Spatial neighborhoods are then constructed (using the definition in \cref{subsec:neighborhood_construction}) for each cell, and provided as input to the frozen NOLAN teacher, which outputs a $K$-dimensional vector of niche-assignment logits per cell. These logits used as the teacher signal for distillation objective (\cref{fig:overview}, Teacher pipeline)

\subsection{Cross-modal distillation from spatial transcriptomics to histology}

We propose a cross-modal distillation approach that transfers transcriptomics-derived niche structure from the teacher to a histology-only student. Intuitively, the teacher encodes expression-defined neighborhood organization (cell state + local context), and distillation transfers this organization into a histology-only student. We train a histology student model to match the teacher’s temperature-softened niche-assignment distribution from histology-derived neighborhood features. We first extract per-cell histology features from a pretrained H\&E foundation model. Specifically, we use UNIv2 \cite{chen2024uni} and for a given H\&E crop $c_i$, we compute $u_i$ = UNIv2($c_i$), using the [CLS] token of the last layer as the cell representation. Subsequently, we construct neighborhood embeddings according to \cref{subsec:neighborhood_construction} and input them to a student transformer which, like the teacher, outputs a $K$-dimensional activation vector (logits): $a_i^{\text{HIST}} \in \mathbb{R}^K$ (\cref{fig:overview}, Student pipeline). We apply a temperature-scaled softmax to obtain softened categorical distributions: $p_{i}^{\text{ST}} = \text{softmax}(a_{i}^{\text{ST}} / \tau)$ and
$p_{i}^{\text{HIST}} = \text{softmax}(a_{i}^{\text{HIST}} / \tau)$, where $\tau$ is the distillation temperature. We minimize the standard distillation loss:
\begin{equation}
\label{eq:distill_loss}
\mathcal{L}_{\mathrm{distill}}
=
\tau^2 \frac{1}{N}\sum_{i=1}^{N}
\mathrm{KL}\!\left(
p_i^{\mathrm{ST}} \,\|\, p_i^{\mathrm{HIST}}
\right),
\end{equation}
where $\mathrm{KL}(\cdot \,\|\, \cdot)$ denotes the Kullback--Leibler divergence.
At inference, the student is applied to histology-only data by computing UNIv2 features, neighborhood embeddings and predicting $a_i^{\text{HIST}}$, from which we derive niche representations (\cref{fig:overview}, At inference pipeline). During training, all upstream encoders (scVI, NOLAN, UNIv2) are frozen; only the student's parameters are optimized.
\section{Experimental evaluation}
\label{sec:evaluation}

\subsection{Datasets}

We use 16 publicly available Xenium datasets from the 10x Genomics public datasets portal \cite{tenx_datasets_portal} spanning 12 tissues: human colon (healthy and cancer) \cite{tenx_xenium_human_colon_preview_2023}, human colorectal (cancer) \cite{tenx_xenium_crc_io_2024}, human liver (healthy) \cite{tenx_xenium_liver_mt_panel_2024}, human lymph node (cancer) \cite{tenx_xenium_prime_preview_lymph_node_2024}, human breast (cancer) \cite{tenx_xenium_prime_breast_cancer_2024, tenx_xenium_breast_biomarkers_s2_middle_2025}, human ovary (cancer) \cite{tenx_xenium_ovarian_cancer_prime}, human brain (cancer) \cite{tenx_xenium_brain_cancer_io_2024}, human cervix (cancer) \cite{tenx_xenium_prime_cervical_cancer_2024}, human kidney (cancer) \cite{tenx_xenium_rcc_gene_protein_2025}, human pancreas (cancer) \cite{tenx_xenium_pdac_io_2024, tenx_xenium_pancreatic_cancer_2023}, human lung (cancer) \cite{tenx_xenium_human_lung_cancer_2024}, mouse colon (healthy) \cite{tenx_xenium_mouse_colon_ff_2024}, whole-mouse (healthy) \cite{tenx_xenium_mouse_pup_2024}. 

For reproducibility, we report histology image resolution in $\mu$m/pixel for each dataset and the per-dataset values during preprocessing. \cite{tenx_xenium_human_colon_preview_2023} (cancer) and \cite{tenx_xenium_pancreatic_cancer_2023} are 0.137 $\mu$m/pixel, and the remaining 14 datasets are 0.274 $\mu$m/pixel. Each dataset provides (i) an H\&E histology image and (ii) cell-resolved spatial transcriptomics measurements, including per cell-gene expression and (iii) spatial coordinates (cell centers) for both modalities.

\paragraph{Data splits}

\begin{figure}[t]
  \centering
  \includegraphics[width=1.0\linewidth]{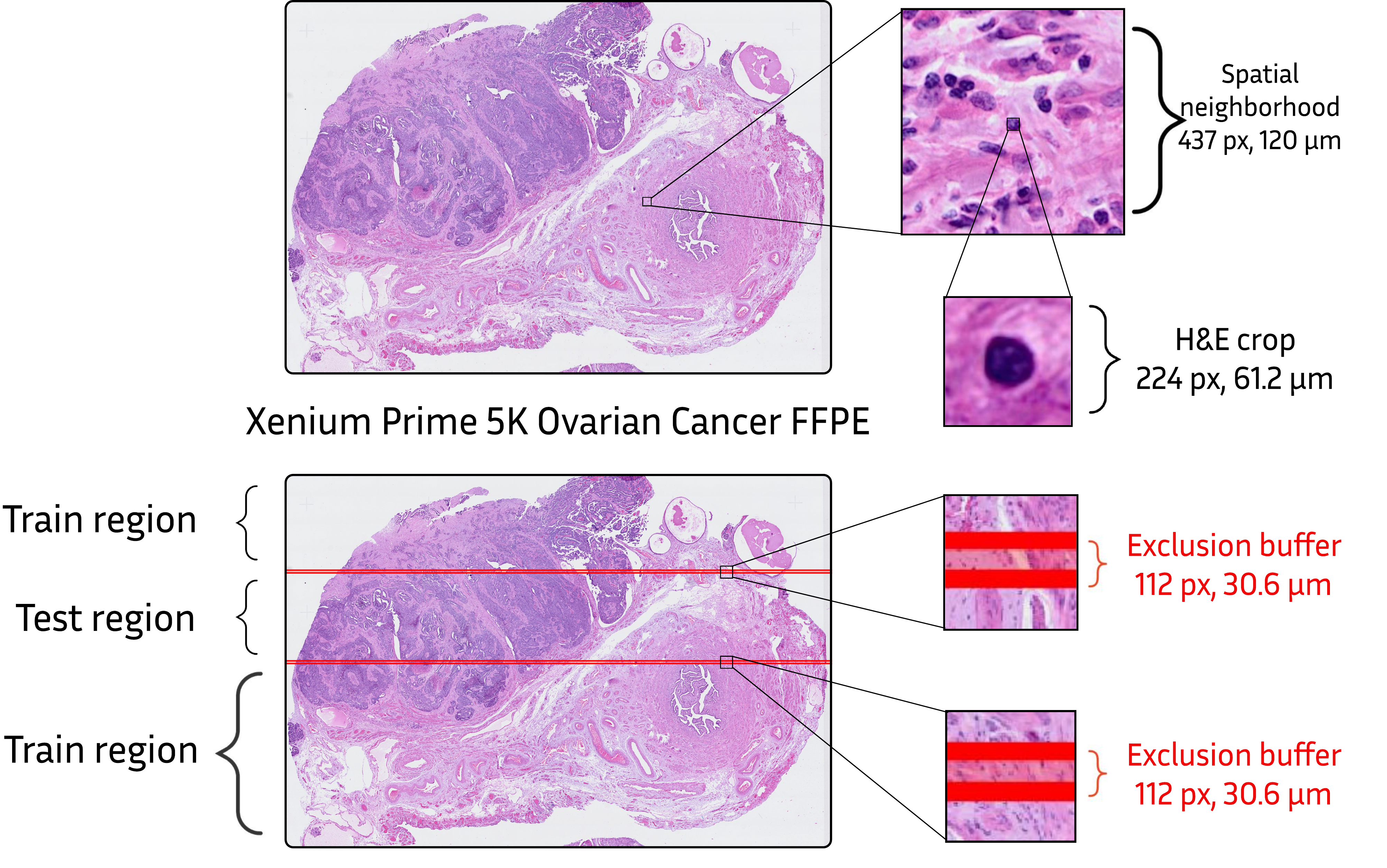}
  \caption{Representative H\&E slide of ovarian cancer. \textbf{Top:} Input image with hierarchical views showing a cellular neighborhood from the tissue and a single-cell crop (the model's basic unit). \textbf{Bottom:} Each tissue is divided into four horizontal continuous strips, where the second-from-top strip is the held-out test data, and the remainder for training. Red bands indicate an exclusion buffer at the test region boundaries. All cells whose centers fall within buffers are discarded and are not used for either training or testing. The buffer size is set to 112 pixels (30.6$\mu$m), which is half of our experimental cell-center crop size of $224 \times 224$ pixels. This design assures that train and test crops are strictly spatially disjoint.}
  \label{fig:input_example}
\end{figure}

Because paired spatial transcriptomics and H\&E datasets remain limited and many public Xenium releases contain only a single paired slide per specimen, our evaluation is performed using within-slide spatial holdout splits. For each slide, we partition the tissue area into four non-overlapping horizontal strips and use the second strip from the top as the test region, with the remaining three strips used for training (\cref{fig:input_example}, bottom). Across datasets, the outermost strips more often contain tissue boundaries, partial tissue coverage, or low cell density, which can make evaluation less representative of the tissue interior. Using the second strip provides a simple and reproducible spatial holdout rule while reducing these edge effects. To prevent train-test leakage, we place an exclusion buffer at each train/test boundary with width equal to half the H\&E crop size. With crops of size $224 \times 224$ pixels, this corresponds to a buffer of 112 pixels (in physical units, this corresponds to 30.6 $\mu$m for slides at 0.274 $\mu$m/pixel and 15.3 $\mu$m for slides at 0.137 $\mu$m/pixel). Cells whose centered crop intersects this buffer are discarded from both training and testing. This ensures that train and test crops are spatially disjoint and that near-boundary context does not leak across splits. Neighborhoods for training cells are constructed using training cells only, and neighborhoods for test cells are constructed using test cells only. The neighborhood radius $r$ is estimated from the training cells and then reused unchanged at test time. All metrics are reported on the held-out test region after exclusion-buffer removal. Because the split is defined spatially on single slides, the resulting cell proportions vary across datasets with tissue geometry and cell density. Across the 16 datasets, there are 8,070,255 cells in total, with per-dataset counts ranging from 190,875 to 1,353,765. Under our spatial holdout protocol, approximately 70\% of cells are assigned to training, 30\% to testing, and 2\% are discarded by the exclusion buffer, with small variation across datasets.

\subsection{Baselines}

We evaluate our method against two histology-only baselines that do not rely on gene expression supervision, in order to quantify the benefit of incorporating transcriptomic guidance. Both baselines use per-cell histology features extracted from the pretrained H\&E foundation model UNIv2 \cite{chen2024uni}: we extract crops of size $224 \times 224$ pixels and apply channel-wise normalization. We define \textbf{Histology-NOLAN} as the first baseline, to show that directly applying NOLAN's framework for histology do not recover the same granularity of niches. In Histology-NOLAN, we train the self-supervised NOLAN framework on the UNIv2 features, followed by an argmax operation to obtain discrete niche labels. We further define \textbf{Histology-Leiden} as the second baseline, to show that a naive clustering might not be sufficient for unsupervised niche-discovery. In Histology-Leiden, we apply Leiden clustering directly to the UNIv2 features and obtain discrete niche labels.  Resolution was tuned on train only to match target $K$ niches, then applied to test.

\subsection{Qualitative assessment}

\begin{figure*}[t]
  \centering  
  \includegraphics[width=1.0\linewidth, height=0.65\textheight, keepaspectratio]{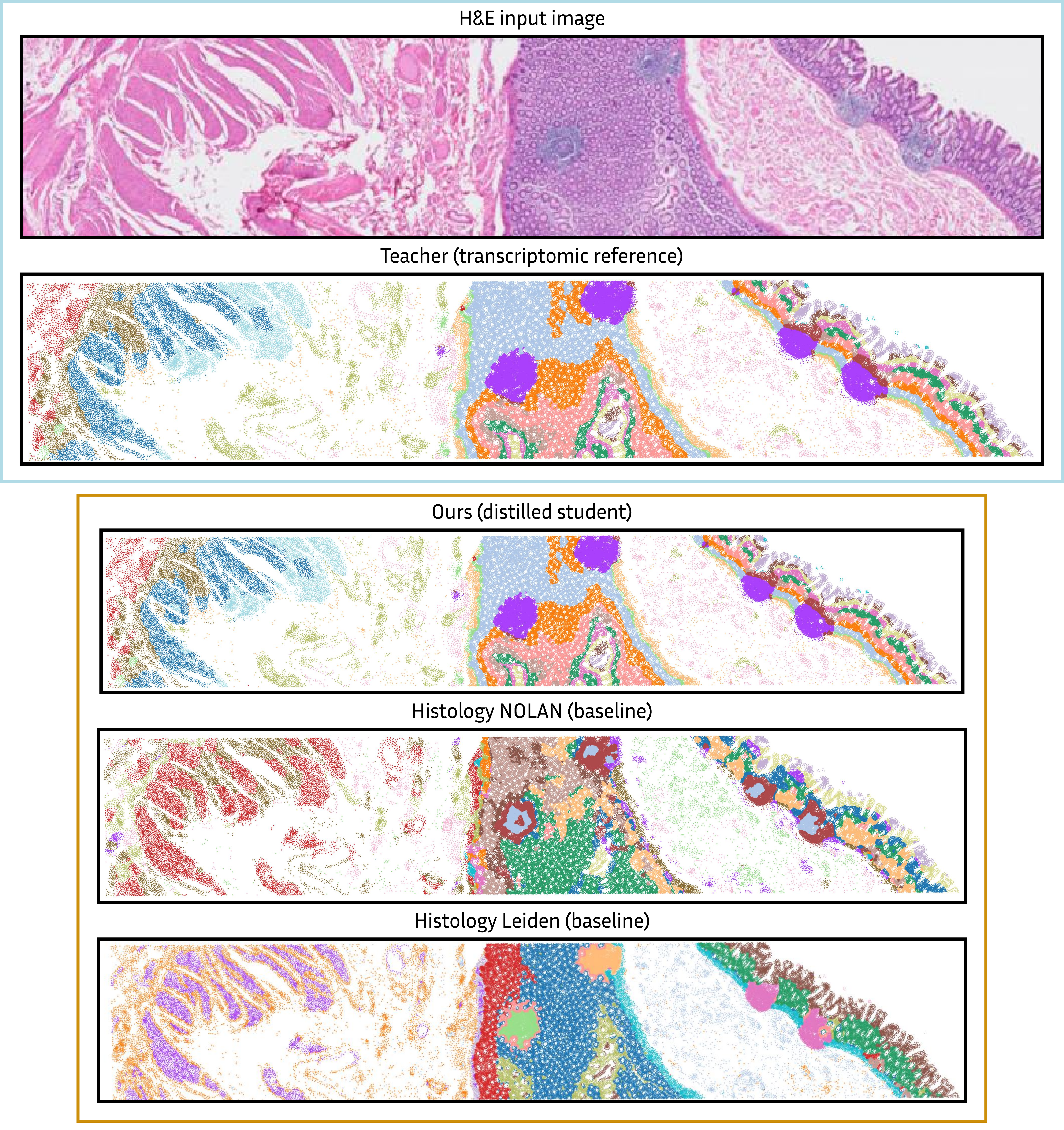}
  \caption{\textbf{Niche segmentation of healthy human colon into 20 distinct niches:} In our supervised distillation framework, the image-based student model (ours) learns to replicate the teacher's expression-derived segmentation using histology features alone. Each color corresponds to a distinct niche. For example, in the teacher and student results, purple patches mark B-cell follicles, warm tones (orange, pink) delineate stromal compartments, and layered greens and blues along the tissue edge reflect epithelial zonation. The student closely reproduces the teacher's spatial organization, preserving both large-scale tissue compartments and fine-grained zonal boundaries. While B-cell follicles are recovered by all methods, subtler structures such as epithelial zonation and stromal layering are resolved only by the teacher and student, highlighting the benefit of transcriptomic supervision. Histology-NOLAN baseline partially recovers coarse compartments but exhibits substantial spatial noise, while Histology-Leiden baseline produces large, spatially uniform blocks that fail to capture the tissue's internal organization.}
  \label{fig:qualitative_results_1}
\end{figure*}

\begin{figure*}[t]
  \centering
  \includegraphics[width=1.0\linewidth, height=0.64\textheight, keepaspectratio]{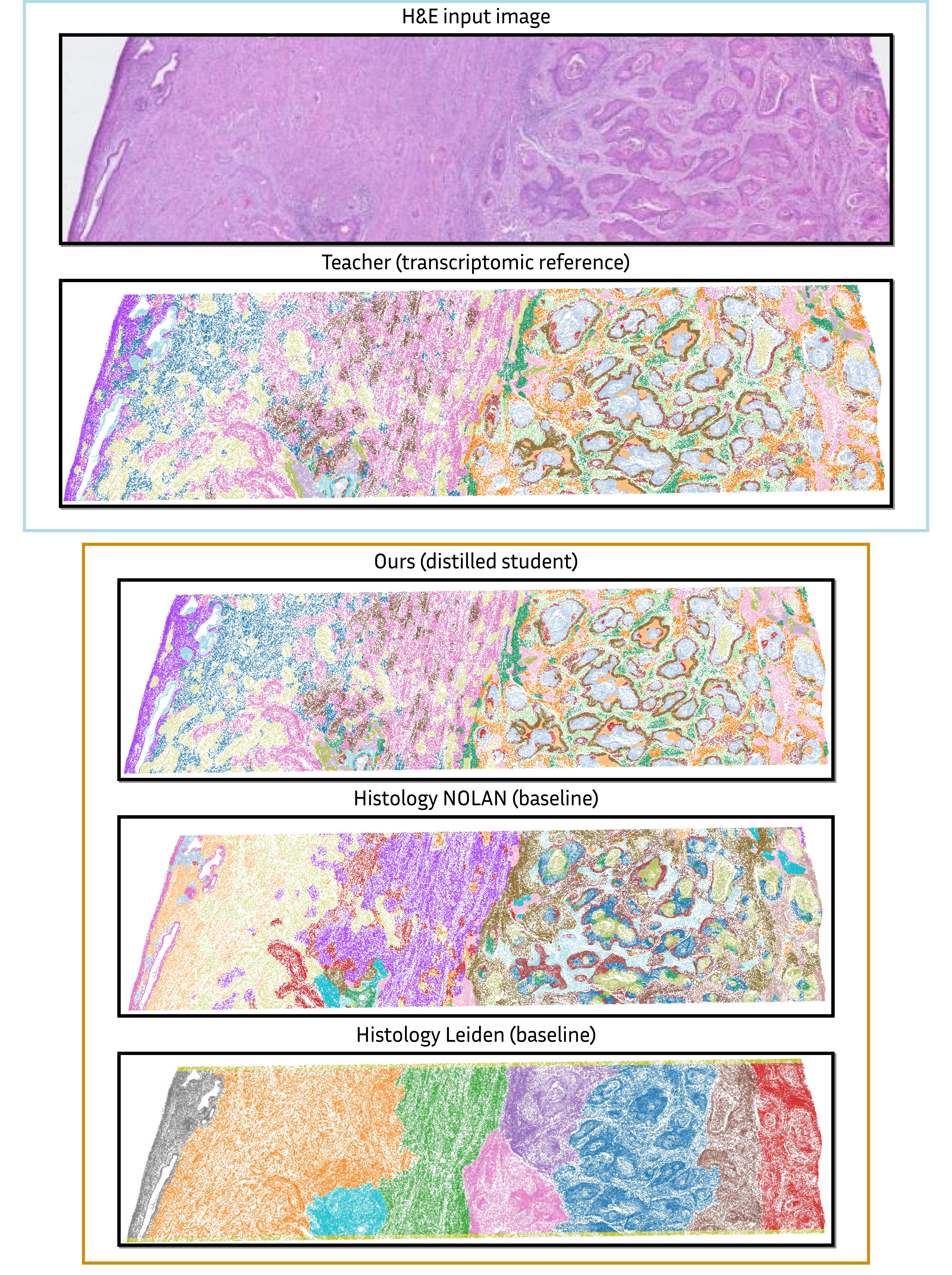}
  \caption{\textbf{Niche segmentation in human cervix cancer into 20 distinct niches}: Our model (student) closely replicates the teacher's expression-derived segmentation from histology alone, with both models recovering fine-grained tissue structure including \textit{tumor epithelial regions}, \textit{stromal infiltration}, and the precise boundaries of \textit{invasive carcinoma nests}. Histology-NOLAN and Histology-Leiden produce substantially coarser segmentation, failing to resolve the spatial heterogeneity captured by Teacher and Student.}
  \label{fig:qualitative_results_2}
\end{figure*}

After performing niche segmentation, we conducted a qualitative assessment of the spatial assignments across all four models for human healthy colon (\cref{fig:qualitative_results_1}) and human cervical cancer (\cref{fig:qualitative_results_2}). Each color in the visualizations corresponds to a distinct niche, and spatially coherent color regions indicate that a model has successfully recovered biologically meaningful tissue structure. Across both tissues, our model (student) consistently reproduces the teacher's spatial organization --- recovering structures such as \textit{B-cell follicles}, \textit{epithelial zonation}, \textit{stromal compartments}, and \textit{invasive carcinoma nests} --- at a resolution that neither unsupervised baseline achieves. Histology-NOLAN partially recovers coarse compartments but produces spatially noisy assignments, while Histology-Leiden segments tissue into large uniform blocks that obscure fine-grained heterogeneity. These observations confirm that transcriptomic supervision enables the student to learn a niche representation that goes substantially beyond what is achievable from histology features alone.

\subsection{Measuring agreement with expression-based tissue segmentation}

Previous work \cite{bakulin2025nolan}, demonstrated that NOLAN, trained on spatial transcriptomic data, identifies biologically meaningful niche boundaries, outperforming methods such as BANKSY, CellCharter, and MENDER. We therefore treat its niche assignments as a gold standard and ask whether comparable quality can be achieved from histology alone at inference time.

To measure \textbf{agreement between niche assignments} of the expression-based segmentation of the tissue and image-based segmentations produced by different methods, we report Adjusted Rand Index (ARI) and Normalized Mutual Information (NMI). We evaluate: (i) agreement between teacher and student niche labels, and (ii) agreement between teacher and each baseline. To assess consistency, we vary the number of niches across $K \in \{10, 20\}$ for each dataset, and report the mean and standard deviation of ARI and NMI across all datasets. As shown in \cref{tab:clustering_performance}, the distilled student consistently attains substantially higher agreement with the spatial-transcriptomics teacher than either unsupervised baseline built on the same histology features. The gains are observed for both ARI and NMI, and remain consistent across $K \in \{10,20\}$, suggesting that cross-modal distillation transfers the teacher-defined niche organization into the histology-only student more faithfully than applying unsupervised clustering directly in the histology feature space.

\subsection{Evaluating consistency cell type composition}

\begin{table}[t]
\centering
\small
\caption{Jensen–Shannon divergence (JSD) between the per-niche cell-type compositions of the teacher and each method $K \in \{10, 20\}$. We compute per-niche cell-type distributions from the published 10x cell-type annotations (evaluation only), align niches to the teacher via one-to-one matching that minimizes JSD, and report the teacher-size-weighted mean JSD. Lower is better; best results are in \textbf{bold}.}
\label{tab:jsd}
\begin{tabular}{lccc}
\toprule
Dataset & Ours & \begin{tabular}[c]{@{}c@{}}Histology \\ NOLAN\end{tabular} & \begin{tabular}[c]{@{}c@{}}Histology \\ Leiden\end{tabular} \\
\midrule
\multicolumn{4}{l}{$K=10$} \\
Human ovary    & \textbf{0.0052} & 0.0557 & 0.0750 \\
Human pancreas & \textbf{0.0042} & 0.0626 & 0.1572 \\
Human breast   & \textbf{0.0024} & 0.0323 & 0.0781 \\
\midrule
\multicolumn{4}{l}{$K=20$} \\
Human ovary    & \textbf{0.0101} & 0.0958 & 0.0851 \\
Human pancreas & \textbf{0.0069} & 0.0622 & 0.1237 \\
Human breast   & \textbf{0.0090} & 0.0244 & 0.0545 \\
\bottomrule
\end{tabular}
\end{table}

Since clustering methods produce arbitrary label permutations, ARI and NMI measure assignment overlap but cannot reveal whether two methods \textbf{discover niches with similar biological identity}. To assess this directly, we analyze the cell-type composition of each niche to measure of whether different methods recover the same underlying tissue zones.
We compute per-niche cell-type compositions using published 10x Genomics annotations, which are available for only three datasets: human ovarian cancer (406K cells) \cite{tenx_xenium_ovarian_cancer_prime}, human pancreatic cancer (190K cells) \cite{tenx_xenium_pancreatic_cancer_2023}, and human breast cancer (380K cells) \cite{tenx_xenium_breast_biomarkers_s2_middle_2025}. These annotations were derived by clustering Xenium profiles and assigning identities using marker-based evidence (Human Protein Atlas, gene set enrichment, literature, and matched single-cell data where available). Crucially, these annotations are withheld from all training, representation learning, and model selection steps, and are used solely for evaluation.
For each model, we treat the per-niche cell-type composition as a probability distribution and measure inter-model distances via Jensen--Shannon divergence (JSD), between that model and the teacher. Since niche labels are not comparable across models, we align niches using a one-to-one assignment that minimizes JSD, and summarize the result as the teacher-weighted mean JSD over paired niches. To confirm this alignment is not spurious, we compare it against 10,000 random one-to-one pairings: across all annotated datasets, baselines, and $K \in \{10, 20\}$, fewer than 0.02\% of random pairings match the alignment-based JSD, confirming the pairing is highly non-trivial. We report JSD between the teacher to all other models for both $K=10$ and $K=20$ in \cref{tab:jsd}. Lower JSD indicates closer agreement in the niche cell-type mixtures, meaning, more similar biological identity of the discovered tissue zones. Across all three annotated datasets, our distilled student achieves the lowest teacher–method JSD for both $K=10$ and $K=20$, while Histology-NOLAN and Leiden are consistently higher. This shows that the student preserves the teacher’s niche-level organization in terms of cell-type composition, beyond label overlap measured by ARI/NMI.

\begin{table}[t]
    \centering
     \caption{Pathology-label probing on the Xenium human ovarian cancer dataset using partial pathology annotations (restricted to Tumor 1, Tumor 2, and Tumor 3; unlabeled cells excluded). We train an SVM to predict pathology labels from method-specific niche assignments, evaluated separately for $K=10$ and $K=20$ niches. Macro-F1 is reported. Best result in each $K$ setting is shown in \textbf{bold}.}
    \small 
    \begin{tabular}{lcc}
        \toprule
               & \multicolumn{2}{c}{Macro-F1} \\ \cmidrule(lr){2-3}
        Method & $K=10$ & $K=20$ \\
        \midrule
        Histology NOLAN & 0.297 & 0.361 \\
        Ours            & \textbf{0.543} & \textbf{0.401} \\
        Teacher         & 0.489 & 0.388 \\
        \bottomrule
    \end{tabular}
    \label{tab:pathology_annotation}
\end{table}

\subsection{Analysis of the alignment with clinical histopathological annotation}

Finally, we evaluate whether the learned niche assignments capture biologically meaningful structure \textbf{independently of NOLAN's expression-based representation}. To this end, we use pathology annotations available for the human ovarian cancer dataset (approximately 230K annotations) \cite{tenx_xenium_ovarian_cancer_prime} as an external reference grounded in clinical tissue characterization. We restrict this analysis to the three most frequent annotated tumor regions (Tumor 1, Tumor 2, and Tumor 3) and train a lightweight SVM to predict pathology labels from the niche assignments of each method, evaluated for $K \in \{10, 20\}$. In both settings, the distilled student achieves higher macro-F1 than the transcriptomic teacher and morphology-only baselines (student $>$ teacher $>$ baseline), as reported in \cref{tab:pathology_annotation}. This suggests that the distilled histology model captures structure that is not only consistent with the teacher, but also better aligned with this manual pathology partition.

\section{Conclusion}
\label{sec:conclusion}

We presented a cross-modal distillation framework that transfers transcriptomics-derived niche structure to a histology-only model. The central premise is that gene expression provides a richer substrate for self-supervised niche learning than morphology alone; we therefore train a self-supervised model on spatial transcriptomic data and use its assignments as a supervision signal for an image-based student, enabling niche segmentation at inference time without transcriptomic profiling. Across multiple Xenium datasets spanning diverse tissues and disease contexts, the distilled student substantially outperforms unsupervised morphology-based baselines in agreement with the transcriptomic teacher, and recovers biologically meaningful niche structure as confirmed by cell-type composition analysis. We believe cross-modal distillation is a promising direction for leveraging the growing body of spatial multi-omics data to improve histological tissue representation.

\section{Limitations}
\label{sec:limitation}

The key limitation of our approach is that: applying the method to a new tissue type requires a paired spatial transcriptomics and histology dataset to train the distillation. Once this bridge dataset is available, the model can be deployed on H\&E data alone; however, the initial transcriptomic pairing cannot be avoided.

\paragraph{Acknowledgments:}
We thank Nathan Levy for his support. This project received funding from the MBZUAI-WIS Joint Program for AI Research, and the Knell Family Institute for Artificial Intelligence.
\newpage
{
    \small
    \bibliographystyle{ieeenat_fullname}
    \bibliography{main}
}

\end{document}